%% file: main.tex
\documentclass[10pt,twocolumn,letterpaper]{article}

\usepackage{cvpr} 
\usepackage{multirow}
\usepackage{float}
\usepackage{placeins}
\usepackage{stfloats}
\usepackage{tabularx}
\newcolumntype{Y}{>{\centering\arraybackslash}X}

\input{preamble}
\definecolor{cvprblue}{rgb}{0.21,0.49,0.74}
\definecolor{Red}{cmyk}{0,1,1,0}
\definecolor{Green}{cmyk}{1,0,1,0}
\definecolor{Cyan}{cmyk}{1,0,0,0}
\definecolor{Purple}{cmyk}{0.45,0.86,0,0}
\definecolor{Rosolic}{cmyk}{0.00,1.00,0.50,0}
\definecolor{Blue}{cmyk}{1.00,1.00,0.00,0}
\definecolor{Orange}{cmyk}{0,0.52,0.80,0}
\definecolor{Black}{cmyk}{1,0,0,1}

\newcommand{\yh}{\textcolor{Black}}

\newcommand{\lm}{\textcolor{Black}}
\newcommand{\fyy}{\textcolor{Black}}
\usepackage[pagebackref,breaklinks,colorlinks,allcolors=cvprblue]{hyperref}
\usepackage{xcolor}  
\newcommand{\lyj}{\textcolor{Black}}

\usepackage{soul}
\sethlcolor{yellow}

\title{\lyj{Unsupervised Structural Scene Decomposition via Foreground-Aware Slot Attention with Pseudo-Mask Guidance}}

\author{Huankun Sheng\\
Tsinghua University\\
{\tt\small shenghk@tsinghua.edu.cn}
\and
Ming Li\\
Tsinghua University\\
{\tt\small mingli\_thu@tsinghua.edu.cn}
\and
Yixiang Wei\\
Tsinghua University\\
{\tt\small  yx-wei22@mails.tsinghua.edu.cn}
\and
Yeying Fan\\
Tsinghua University\\
{\tt\small  fyydemail@gmail.com}
\and
Yu-Hui Wen\\
Beijing Jiaotong University\\
{\tt\small  yhwen1@bjtu.edu.cn}
\and
Tieliang Gong\\
Xi'an Jiaotong University\\
{\tt\small  adidasgtl@gmail.com}
\and
Yong-Jin Liu\\
Tsinghua University\\
{\tt\small  liuyongjin@tsinghua.edu.cn}
}

\begin{document}
\maketitle
\input{sec/0_abstract}    
\input{sec/1_intro}
\input{sec/2_related_work}

\input{sec/3_method}
\input{sec/4_experiment}

\input{sec/5_conclusion}
\input{sec/X_suppl}
{
    \small
    \bibliographystyle{unsrt}
    \bibliography{main}
}


\end{document}

%% file: sec/0_abstract.tex
\begin{abstract}
\lyj{Recent advances in object-centric representation learning have shown that slot attention–based methods can 
effectively decompose visual scenes into object slot representations without supervision. However, existing approaches typically process foreground and background regions indiscriminately, often resulting in background interference and suboptimal instance discovery performance on real-world data. To address this limitation, we propose Foreground-Aware Slot Attention (FASA), a two-stage framework that explicitly separates foreground from background to enable precise object discovery. In the first stage, FASA performs \yh{a} coarse scene decomposition to distinguish foreground from background regions through a dual-slot competition mechanism. These slots are initialized via a clustering-based strategy, yielding well-structured representations of salient regions. In the second stage, we introduce a masked slot attention mechanism where the first slot captures the background while the remaining slots compete to represent individual foreground objects. To further address over-segmentation of foreground objects, we incorporate pseudo-mask guidance \yh{derived from a patch affinity graph constructed with self-supervised image features} to guide the learning of foreground slots. Extensive experiments on both synthetic and real-world datasets demonstrate that FASA consistently outperforms state-of-the-art methods, validating the effectiveness of explicit foreground modeling and pseudo-mask guidance for robust scene decomposition and object-coherent representation. Code will be made publicly available.}
\end{abstract}

%% file: sec/1_intro.tex
\section{Introduction}
\label{sec:intro}

\lyj{A fundamental objective in computer vision and representation learning is to interpret visual scenes by identifying and understanding their constituent objects \cite{singh2025glass, elsayed2022savi++}. Object-centric learning (OCL) addresses this goal by decomposing an image into a collection of object-specific latent representations, thereby enabling structured and interpretable scene understanding \cite{locatello2020object, aydemir2023self}. Such representations benefit various downstream tasks, including object localization \cite{simeoni2023localization} and zero-shot object discovery \cite{singh2025glass}.}


\lyj{Among existing OCL approaches, slot attention (SA) \cite{locatello2020object, tian2025pay, kakogeorgiou2024spot, seitzer2022dinosaur} has emerged as a prominent framework for unsupervised object-centric representation learning. It iteratively groups visual features into a set of latent slots, each intended to capture a distinct object in the scene. Early SA methods \fyy{process} the entire image 
\yh{in a holistic manner,} without explicitly distinguishing foreground from background regions \cite{locatello2020object}. As a result, in complex real-world images, background textures and illumination variations can distract the attention mechanism, leading slots to capture irrelevant background content rather than foreground objects.}

\lyj{Recent efforts have sought to incorporate implicit cues that encourage slot attention to focus more strongly on foreground regions \cite{tian2025pay}. While beneficial, these methods only provide implicit, feature-level guidance and lack explicit structural separation between foreground and background, leaving models vulnerable to background interference.}


\lyj{Furthermore, slot representations learned by existing approaches often use a fixed number of slots and suffer from over-segmentation, where a single object is incorrectly split across multiple slots. To mitigate this, GLASS \cite{singh2025glass} introduces semantic guidance using cross-attention maps from a pre-trained diffusion model \cite{rombach2022imagegen}, which helps align each slot with a complete object region via semantic masks. However, generating these masks requires image captions produced by vision–language models \cite{li2023blip}, complicating the pipeline and making mask quality highly dependent on the accuracy of generated text descriptions.}


\lyj{To overcome these limitations, we propose {\bf Foreground-Aware Slot Attention (FASA)}, a novel approach that enhances scene decomposition by explicitly focusing on foreground objects while alleviating over-segmentation. FASA first separates the scene into foreground and background components, then decomposes the foreground region into distinct object representations. For foreground–background separation, we introduce two dedicated slots that aggregate features extracted by a Vision Transformer (ViT) \cite{dosovitskiy2020image}. Unlike previous methods that initialize slots randomly, FASA employs a clustering-based initialization strategy, which yields more stable and semantically meaningful slot representations and improves decomposition accuracy. Based on this separation, we design a masked slot attention module where the first slot models the background and the remaining slots capture individual foreground objects. This design enables instance-level segmentation by binding each foreground object to a distinct slot.}

\lyj{Additionally, to optimize the allocation of a fixed number of slots and prevent over-segmentation—especially when the number of foreground instances is smaller than the number of foreground slots—we leverage MaskCut \cite{wang2023cutler} to generate pseudo-masks that guide slot learning. \yh{Specifically, these masks are obtained by applying Normalized Cut~\cite{wang2023tokencut} via a fully connected patch-affinity graph, constructed by computing cosine similarities between self-supervised DINO  \cite{caron2021dino} features extracted from each image patch using a ViT.} By incorporating regional information from these masks, slot attention is encouraged to assign patches belonging to the same object to a single slot, thereby preserving object integrity and reducing fragmentation.}

\lyj{In summary, our main contributions are as follows:
\begin{itemize}
   \item 
   We propose a hierarchical slot attention mechanism that first disentangles foreground and background regions via clustering-based initialization, then refines foreground object decomposition using a masked slot attention strategy. This two-stage design structures representation learning from coarse to fine, enabling robust and precise scene decomposition.
   \item
   \yh{
   We introduce robust pseudo masks, derived from a patch-affinity graph of self-supervised visual features, to guide slot representation learning,  thereby establishing an effective mechanism to address over-segmentation.}
   \item 
   Experimental results on both synthetic and real-world datasets demonstrate that our method achieves state-of-the-art performance and exhibits 
   \yh{robust} generalization on downstream tasks such as object localization and zero-shot object discovery.
\end{itemize}}

%% file: sec/2_related_work.tex
\section{Related Work}
\label{sec:formatting}

\subsection{Unsupervised Object Detection}

\lyj{Recent advances in self-supervised Vision Transformers (ViTs) \cite{dosovitskiy2020image} have demonstrated significant potential for uncovering semantic \yh{image} structures 
without human annotation. Notably, DINO revealed 
 that self-supervised ViTs can learn to perform latent semantic segmentation—a capability 
\yh{absent} in their supervised counterparts \cite{caron2021dino}. Inspired by this finding, LOST \cite{simeoni2021lost} 
\yh{identified} foreground regions without cross-image matching or external supervision, achieving strong performance in single-object detection. TokenCut \cite{wang2023tokencut} further 
\yh{extended} this paradigm by formulating object detection as a graph partitioning problem. 
\yh{It applied normalized cuts to transformer features within a unified, training-free framework to segment foreground objects in both images and videos, demonstrating performance that substantially outperformed prior approaches.}}


\lyj{Beyond single-object detection, FreeSOLO \cite{wang2022freesolo} introduced  a contrastive self-supervised framework for unsupervised multi-object discovery. Building upon the SOLO architecture \cite{wang2021solo}, it 
\yh{employed} a two-stage pipeline: 
\yh{a Free Mask module first \fyy{extracted} coarse object masks, which were then refined through weak supervision and self-training in a subsequent self-supervised SOLO stage}. Similarly, MaskDistill \cite{van2022maskdistill} directly 
\yh{distilled} object masks from transformer attention maps, \yh{by} clustering them into pseudo–ground truth annotations and refining the results through confidence filtering. CutLER \cite{wang2023cutler} proposed a simple yet effective Cut-and-Learn framework for unsupervised object detection and instance segmentation. It 
\yh{harnessed} the inherent object localization capability of self-supervised transformers through MaskCut \cite{wang2023cutler}, an algorithm that 
\yh{generated} multi-object pseudo-masks, followed by iterative self-training for refinement.}


\lyj{U2Seg \cite{niu2024u2seg} unified 
this evolving field by introducing the first unsupervised universal segmentation framework, capable of performing semantic, instance, and panoptic segmentation within a single model. In parallel, the contrastive saliency network (CSNet)  \cite{guan2025contrastive} 
\yh{explored} unsupervised salient object detection from a representation-learning perspective. Rather than relying on traditional low-level priors such as color or boundary contrast, CSNet 
\yh{introduced} a contrastive learning framework that discovers saliency through instance discrimination and feature re-coordination, \yh{thereby} enabling robust detection of visually distinctive regions without supervision.}


\subsection{Object-Centric Learning}

\lyj{Object-centric learning aims to decompose visual scenes into structured, object-level representations. Early approaches such as MONet \cite{burgess2019monet} and IODINE \cite{greff2019IODINE} adopted a spatial mixture model paradigm, leveraging sequential attention or iterative variational inference to recover object structure. GENESIS \cite{engelcke2019genesis} 
\yh{further advanced this line of work} by incorporating explicit modeling of latent spatial layouts. However, these methods often depend\yh{ed} on recurrent processing or variational optimization, which 
\yh{limited} their efficiency and scalability.}


\begin{figure*}[htbp]
\setlength{\abovecaptionskip}{0.08cm}
    \centering
    \includegraphics[width=0.9\textwidth]{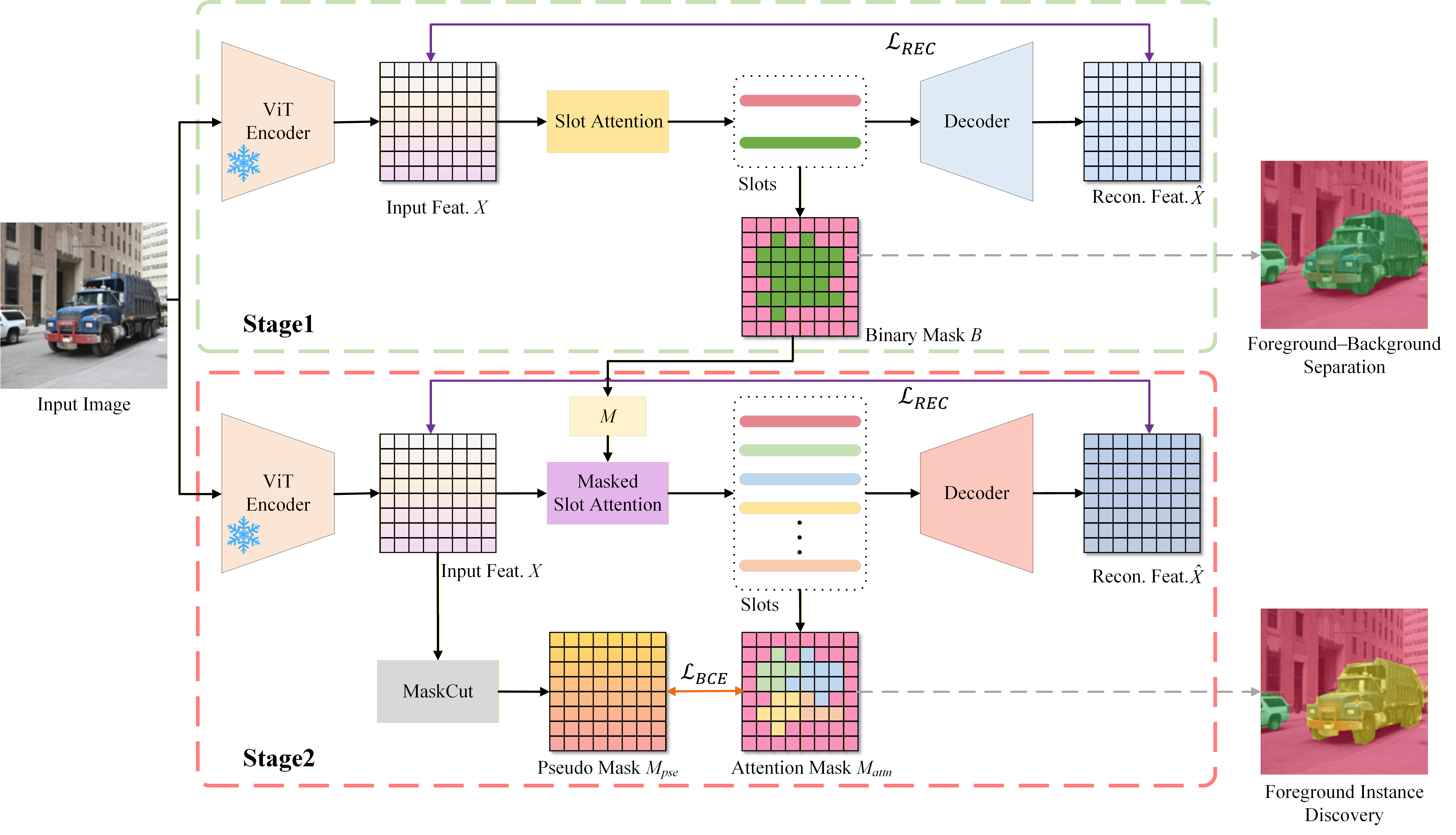}
    \caption{Architecture of the Foreground-Aware Slot Attention (FASA). \lyj{Our framework operates in two sequential stages. First, a two-slot attention module, trained with feature reconstruction loss, generates a binary mask separating foreground from background regions.} Conditioned on this mask, the second stage introduces a masked slot attention mechanism that binds input features to different slots: the first slot is dedicated to representing the background, while the remaining slots correspond to foreground objects. In addition, pseudo labels obtained from MaskCut \cite{wang2023cutler} are used to guide the learning of the foreground slots.}
    \label{fig:architecture}
\end{figure*}

\lyj{In contrast, {\bf slot attention} \cite{locatello2020object} introduced a simple, differentiable, and end-to-end framework that iteratively binds visual features to a set of latent slots, each intended to capture a distinct object entity. This mechanism has become a foundational component in many subsequent object-centric models \cite{wang2023slotVAE, zhang2024gated, lee2024guided, lsd, wu2023slotdiffusion}. Despite its efficiency, traditional slot attention often 
\yh{failed} to form coherent object representations in complex real-world scenes. 
\yh{This limitation stemmed  primarily from its reliance on pixel-space reconstruction, which provided  suboptimal supervisory signals that emphasized  low-level visual cues—such as color and texture—at the expense of higher-level semantic content.}}


\lyj{To overcome this limitation, recent methods have explored more semantically meaningful training objectives and architectural improvements. DINOSAUR \cite{seitzer2022dinosaur} advanced real-world object-centric learning by reconstructing self-supervised features rather than raw pixels. Similarly, SLATE \cite{singh2021slate} introduced a transformer-based autoregressive decoder conditioned on object slots, 
\yh{which enabled} richer spatial and semantic modeling, while SPOT~\cite{kakogeorgiou2024spot} improved training stability through self-training and patch-order permutation strategies, achieving stronger performance on real-world benchmarks. Parallel efforts such as Adaptive Slot Attention \cite{fan2024adaptive} and DIAS \cite{zhao2025dias} refined the slot aggregation mechanism through slot selection or slot re-initialization and self-distillation, leading to improved slot utilization and convergence behavior. GLASS \cite{singh2025glass} further introduced latent slot diffusion guided by semantic priors to alleviate over- and under-segmentation, though it 
\yh{relied} on 
\yh{the accuracy of text descriptions.}}

\yh{The limitation of the aforementioned methods \fyy{lies} in their lack of effective separation of foreground and background, which limited their ability to handle complex real-world scenes, as the reconstruction objectives were easily dominated by low-level visual cues such as color and texture rather than object-level semantics. To address this, FB-Indicator \cite{tian2025pay} integrated a contrastive foreground–background indicator that injected semantic guidance into slot allocation, reducing background interference and improving segmentation quality on both synthetic and real datasets.  However, this approach still suffered from ambiguous slot assignments and relied on post-processing to correct over-segmentation.}

%% file: sec/3_method.tex
\section{Method}

\lyj{This section introduces the proposed FASA model. As illustrated in Figure~\ref{fig:architecture}, our approach builds on the slot attention mechanism to enable structured scene decomposition through foreground–background separation and pseudo-mask guidance. We begin by reviewing the fundamentals of slot attention in Section~\ref{subsec:pre}. Section~\ref{subsec:fb} details our hierarchical slot attention design for foreground–background disentanglement. Section~\ref{subsec:Pseudo} describes the generation of pseudo-masks for supervision, and Section~\ref{subsec:loss} presents the overall loss function used for training.}

\subsection{Preliminary on Slot Attention}
\label{subsec:pre}

Given \lyj{an initial set of} slots $S^0 \in \mathbb{R}^{K \times D_{\text{slots}}}$—typically sampled from a Gaussian distribution—slot attention iteratively refines these $K$ latent vectors to aggregate \yh{information from} the visual tokens 
$X \in \mathbb{R}^{N \times D_{\text{inputs}}}$, where each slot strives to explain a subset of tokens. For \yh{the} $t$-th refinement 
\yh{iteration,} the token-to-slot attention is computed as:
\begin{equation}
\small
\text{logits} = \frac{k(X)q(S)^T}{\sqrt{D}}, \quad A = \text{softmax}(\text{logits}) \in \mathbb{R}^{N \times K}
\label{eq:logits}
\end{equation}
where $k(\cdot)$, $q(\cdot)$ are learnable linear transformations 
\yh{that} map $X$ and $S$ to a common dimension $D$, and softmax$(\cdot)$ is normalized over the slot dimension for each token (row-wise) to encourage competition. 
\yh{Subsequently,  the input values are aggregated via a weighted average to derive the updates for each slot:} 
\begin{equation}
U = W^T v(X) \in \mathbb{R}^{K \times D}, \quad W_{ij} = \frac{A_{ij}}{\sum_{i} A_{ij}}
\label{eq:U}
\end{equation}
where $v(\cdot)$ is a linear transformation \yh{applied to $X$}. Each slot is then updated by a GRU \cite{cho2014gru} followed by 
\yh{a multi-layer perceptron (MLP) \cite{watters2019MLP}} with \yh{a} residual \yh{connection}:
\begin{equation}
S \leftarrow \text{GRU}(S, U), \quad S \leftarrow S + \text{MLP}(\text{LN}(S))
\label{eq:S}
\end{equation}
The updated set of slots is then fed into a decoder to reconstruct the input. The decoder can be implemented as a simple 
MLP \cite{watters2019MLP}, a transformer-based model \cite{singh2022TRANSFORMERDECODER}, or a diffusion-based decoder \cite{wu2023slotdiffusion}.

\subsection{Foreground-Aware Slot Attention}
\label{subsec:fb}

\lyj{To obtain semantically coherent object representations, the proposed FASA framework first explicitly separates foreground and background regions. This initial stage aims to disentangle salient foreground areas containing primary objects from less informative background context.}

As shown in Figure~\ref{fig:architecture}, given the feature map extracted 
\yh{by} the encoder, we use two distinct slots to segment the foreground and background \lyj{within} each image. \lyj{While prior methods \cite{locatello2020object, kakogeorgiou2024spot, tian2025pay} typically initialize slots by sampling from a Gaussian distribution $\mathcal{N}(\mu, \text{diag}(\sigma^2))$, we observe that such random initialization often leads to insufficient foreground-background separation after training. Inspired by findings from \cite{wang2023tokencut}, which demonstrate that ViT features encode semantically rich information suitable for segmentation, we introduce a clustering-based initialization strategy, where the cluster centers derived from ViT features $X$ are used to initialize the slots.}
\yh{Specifically, we apply K-Means++ seeding \cite{arthur2006kmeans++} on $X$ to obtain two centroids $\mu_1, \mu_2 \in \mathbb{R}^D$, which are then mapped to the slot space via a linear transformation to form the two initial slots.} 

After training, the two slots specialize in representing the foreground and background of the image, respectively. A binary mask is then constructed from the attention map to partition the image into two disjoint sets of patches. To 
\yh{identify} which set corresponds to the foreground, we employ a simple yet empirically effective object-centric prior~\cite{maji2011prior}
, which states that the foreground region contains fewer than two of the four image corners. If this criterion is \lyj{violated}, the assignments of the foreground and background partitions are \lm{swapped}. The resulting binary mask can be denoted as:
\begin{equation}
B = \{b_i\}_{i=1}^N, \quad b_i \in \{0, 1\}
\label{eq:B}
\end{equation}
where $b_i = 1$ represents the $i$-th patch belongs to the foreground, and $b_i = 0$ denotes the background. \lm{This} separation \lyj{establishes} a structured foundation for subsequent slot-based decomposition \lm{over} object-relevant regions.

\yh{Building on} the separation of foreground and background, \lyj{our} FASA framework 
\yh{proceeds to} \lyj{discover and decompose individual objects within the foreground region.} \lm{In this stage, a new set of learnable slots iteratively attend to distinct objects,}
enabling unsupervised instance-level decomposition.

For foreground decomposition, a straightforward approach is to apply slot attention directly to the foreground patches. However, since the number of foreground patches varies across images, \lm{this results in inconsistent input sizes and unstable training.} 
To address \lm{this,}
we design a masked slot attention module conditioned on the foreground–background separation results.  This module explicitly assigns the first slot to represent the background, while the remaining slots \lm{model individual foreground objects,}
thereby enabling structured scene decomposition.

Specifically, we construct a mask $M \in \mathbb{R}^{N \times K}$ based on ${B}$.  The value of the first column in $M$ is set as:

\begin{equation}
M_{i1} = 
\begin{cases} 
+\infty, & \text{if } b_i = 0, \\
-\infty, & \text{if } b_i = 1 
\end{cases}
\label{eq:M}
\end{equation}
where $\infty$ \lm{denotes} a large value, e.g., $10^6$, and all other columns of $M$ are \lm{set to} zero. We then add $M$ to the attention logits in Eq. (\ref{eq:logits}). After applying the softmax operation along the slot dimension, this design \lm{enforces the first slot to consistently represent}
the background, while the remaining slots compete to represent different foreground objects.

\subsection{Pseudo Mask Generation}
\label{subsec:Pseudo}
The \lm{learned} slot representations often suffer from the problems of over-segmentation \cite{singh2025glass}, where patches belonging to the same object are assigned to different slots. To \lm{address} this issue, we employ pseudo masks to guide the learning of slot representations. 
\yh{Specifically, we generate pseudo masks $M_{\text{pse}}$ using the unsupervised MaskCut method \cite{wang2023cutler}.}

MaskCut discovers multiple objects \yh{in an image} by iteratively applying Normalized Cuts (NCut) \cite{wang2023tokencut} on a patch-affinity graph built from self-supervised ViT features. Given an image, it extracts DINO ``key'' features for each patch and constructs a fully connected graph with cosine similarities \yh{between these features}:
\begin{equation}
W_{ij} = \frac{K_i K_j^T}{\|K_i\| \|K_j\|}
\label{eq:W}
\end{equation}
where $K_i$ is the ``key'' feature of patch $i$. NCut then solves \lyj{the equation} $(D - W)x = \lambda Dx$ to obtain a bipartition from the second smallest eigenvector, where $D$ is a $N \times N$ diagonal matrix with $d(i) = \sum_j W_{ij}$. The foreground is then determined from this bipartition using the simple prior~\cite{maji2011prior}, as mentioned in Section~\ref{subsec:fb}. The discovered foreground patches are then masked out, and 
NCut is reapplied to the masked graph to extract additional object instances. 
\yh{This process typically repeats for three iterations. Full details are provided in the supplementary material.}

Once the pseudo ground-truth masks $M_{\text{pse}}$ are generated, 
\lm{they are used} to guide the slots. 
\yh{ Specifically, we} extract the attention masks $M_{\text{attn}}$ for each slot using the attention matrix from Eq.~(\ref{eq:logits}). 
\yh{Each attention mask is then matched to a corresponding component in the pseudo masks $M_{\text{pse}}$, which we formulate as a bipartite matching problem and solve using the Hungarian algorithm \cite{kuhn1955hungarian}.} 
\yh{Formally, give} $O$ slots with attention masks and pseudo masks $M_{\text{pse}}$ containing $F$ segments, we compute a binary assignment matrix $P \in \{0, 1\}^{O \times F}$ 
\yh{as follows:}
\begin{equation}
\arg \min_P \sum_{i=1}^O \sum_{j=1}^F -c_{ij} p_{ij}
\label{eq:matching}
\end{equation}
where $p_{ij} \in \{0, 1\}$ indicates whether slot $i$ is matched to segment $j$. The optimization enforces a one-to-one assignment for each slot. The cost $c_{ij}$ is defined as the mean Intersection-over-Union (IoU) between the predicted mask of slot $i$ and segment $j$ in the pseudo masks $M_{\text{pse}}$.

\lm{Guided by the mask generated by MaskCut,}
slot attention utilizes the regional information from the mask to assign patches belonging to the same object to a single slot, thereby preventing object fragmentation into multiple slots and effectively mitigating the over-segmentation issue.

\subsection{Loss Function}
\label{subsec:loss}
\lyj{In the foreground–background separation stage}, we \lyj{adopt a} feature reconstruction loss following \cite{seitzer2022dinosaur}. Let $X, \hat{X} \in \mathbb{R}^{N \times D_{\text{inputs}}}$ denote the ground-truth features and their reconstructions. Training minimizes the reconstruction objective:
\begin{equation}
L_{\text{REC}} = \frac{1}{N D_{\text{inputs}}} \| X - \hat{X} \|_2^2
\label{eq:rec}
\end{equation}
\lyj{In the} foreground decomposition stage, we optimize \lyj{a combined objective consisting of the} feature reconstruction loss $L_{\text{REC}}$ and a binary cross-entropy loss $L_{\text{BCE}}$ between the pseudo mask $M_{\text{pse}}$ and the attention mask from the slots $M_{\text{attn}}$. The binary cross-entropy loss is only computed on the matched slots, according to the matching matrix $P \equiv P(M_{\text{pse}}, M_{\text{attn}})$. The full loss is given by:
\begin{equation}
L = L_{\text{REC}} + \lambda L_{\text{BCE}}(P(M_{\text{pse}}, M_{\text{attn}}))
\label{eq:loss}
\end{equation}

%% file: sec/4_experiment.tex
\section{Experiment}

\subsection{Setup}

\paragraph{Datasets.}
We evaluate our method on \lyj{both} synthetic and real-world datasets. For synthetic evaluation, we adopt the MOVi-C \cite{greff2022movic}, which contains approximately 1,000 realistic 3D-scanned objects \lyj{distributed across scenes with 3–10 objects each}. 
\yh{Following the practice in \cite{seitzer2022dinosaur}, we randomly sample frames from this video-based dataset for our experiments.} For real-world evaluation, we use MS COCO 2017 \cite{lin2014coco}, which contains diverse natural images with multiple co-occurring objects, making it a challenging benchmark for object-centric methods. We also include PASCAL VOC 2012 \cite{everingham2010voc}, which typically features one or a few dominant objects per image and therefore provides a more straightforward evaluation setting. In addition, we evaluate the zero-shot generalization capability of our model using the CLEVRTex \cite{karazija2021clevrtex} and Obj365 \cite{shao2019objects365} datasets.

\paragraph{Metrics.}
We evaluate object-centric learning performance using Mean Best Overlap (mBO) and Mean Intersection-over-Union (mIoU). The mBO metric is computed at two levels—instance-level (mBO\(^i\)) and category-level (mBO\(^c\))—by pairing each predicted mask with the 
\yh{ground-truth} instance or semantic mask that yields the highest overlap. For mIoU, the predicted masks are first aligned with the ground-truth annotations using the Hungarian assignment algorithm to ensure optimal one-to-one matching before computing the intersection-over-union scores.

\paragraph{Implementation Details.}
We train 
\yh{our} model using the Adam optimizer with parameters $\beta_1 = 0.9$, $\beta_2 = 0.999$, no weight decay, and a batch size of 64. The initial learning rate is set to 0.001, following a linear warm-up for the first 10,000 optimization steps and a cosine decay schedule thereafter. 
\yh{The model is trained in two stages, with COCO and PASCAL trained for 100 and 300 epochs, and MOVi-C for 50 and 100 epochs, respectively. } We adopt a ViT-S/14 encoder initialized with DINOv2 \cite{oquab2023dinov2} and 
an MLP decoder. Unless otherwise specified, the binary cross-entropy (BCE) loss is weighted by $\lambda = 0.05$. Following \cite{kakogeorgiou2024spot}, the number of slots is set to 7, 6, 11 for COCO, PASCAL, MOVi-C, respectively. All experiments are conducted on a single RTX 3090Ti GPU. 
\yh{Additional implementation details are provided in the supplementary material.}

\begin{figure*}[htbp]
    \centering
    \includegraphics[width=\textwidth]{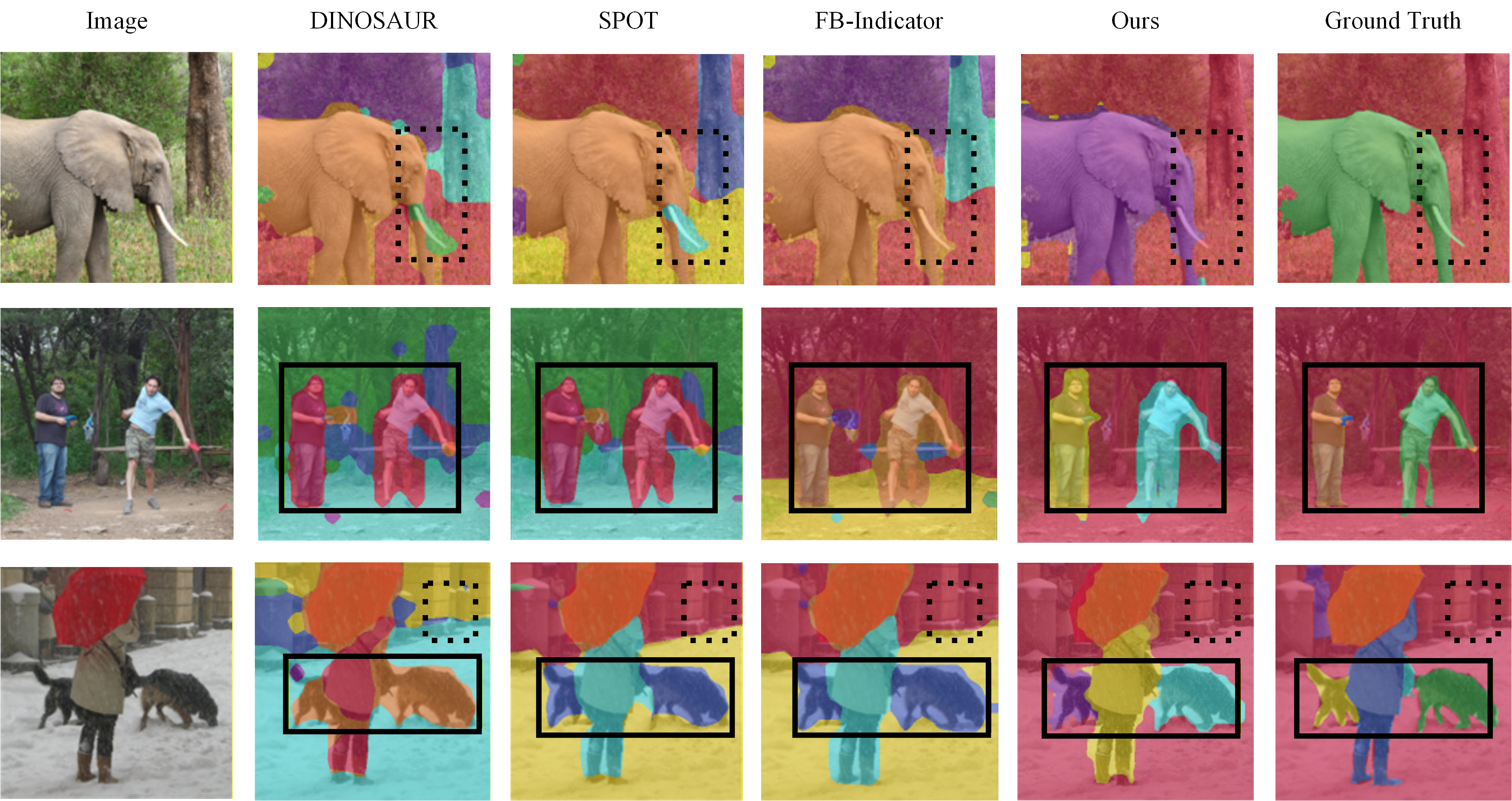}
    \caption{
    \yh{Qualitative comparisons} \lyj{of} object discovery. \lyj{Existing} methods tend to over-segment both background regions and foreground objects, as indicated by the dashed boxes. \lyj{They also struggle to achieve precise instance-level decomposition,} as highlighted by the solid boxes. \lyj{In comparison, our method produces object masks that align closely with the ground-truth annotations.}}
    \label{fig:2}
\end{figure*}

\subsection{Comparison with Object-Centric Methods}
Table \ref{tab:comparison} summarizes the \lyj{quantitative} performance of \lyj{different methods} on the COCO, PASCAL, and MOVi-C benchmarks. \lyj{Our method consistently outperforms representative baselines \cite{locatello2020object,singh2021slate,seitzer2022dinosaur,lsd,kakogeorgiou2024spot,zhao2025dias,tian2025pay}  in terms of the mIoU and instance-level  mBO metrics across all datasets, demonstrating that explicit foreground decomposition combined with pseudo-mask guidance enhances both object discovery and segmentation accuracy.}

Figure~\ref{fig:2} \lyj{provides} qualitative comparisons of object discovery results. \lyj{Our method shows} superior capability in both scene decomposition and foreground object discovery, \lyj{effectively separating foreground objects from background regions to produce accurate and structurally coherent scene representations. Furthermore, it exhibits enhanced robustness in identifying salient objects while suppressing interference from cluttered backgrounds in an instance-centric manner, enabling precise object localization even in visually complex environments.} 

\begin{table}[!t]
\setlength{\abovecaptionskip}{0.1cm}
\centering
\caption{Comparison of \lyj{different methods} on MOVi-C, COCO, and VOC datasets.}
 \resizebox{\columnwidth}{!}{
\begin{tabular}{l|cc|ccc|ccc}
\toprule
\multirow{2}{*}{Model} & 
\multicolumn{2}{c|}{MOVi-C} & 
\multicolumn{3}{c|}{COCO} & 
\multicolumn{3}{c}{VOC} \\ 
 & mIoU & mBO$^i$ & mIoU & mBO$^i$ & mBO$^c$ & mIoU & mBO$^i$ & mBO$^c$ \\ 
\hline
SA \cite{locatello2020object} & -- & 26.3 & -- & 17.2 & 19.2 & -- & 24.6 & 24.9 \\
SLATE \cite{singh2021slate} & 37.8 & 39.4 & -- & 29.1 & 33.6 & -- & 35.9 & 41.5 \\
DINOSAUR \cite{seitzer2022dinosaur} & 41.8 & 42.4 & 31.6 & 33.3 & 41.2 & 42.0 & 43.2 & 47.8 \\
StableLSD \cite{lsd} & 44.2 & 45.6 & 24.7 & 25.9 & 30.0 & 31.5 & 32.1 & 35.4 \\
SPOT \cite{kakogeorgiou2024spot} & 46.4 & 47.0 & 33.0 & 35.0 & 44.7 & 48.8 & 48.3 & 55.6 \\
DIAS \cite{zhao2025dias} & -- & -- & 30.1 & 32.8 & -- & 42.8 & 44.8 & -- \\
FB-Indicator \cite{tian2025pay} & 47.8 & 49.0 & -- & 35.7 & \textbf{45.3}& -- & 49.3 & 56.5 \\
\textbf{Ours} & \textbf{48.2} & \textbf{49.5} & \textbf{34.1} & \textbf{36.5} & 43.9& \textbf{49.5}& \textbf{50.2}& \textbf{57.3}\\
\bottomrule
\end{tabular}
}
\label{tab:comparison}
\end{table}

\lyj{In contrast, competing methods reveal notable limitations when processing challenging scenes. They frequently fragment background regions into multiple disjoint segments and suffer from under-segmentation, where distinct objects are incorrectly merged into single regions. Additional results are provided in the supplementary material.}



\subsection{\lyj{Discussion and} Ablation Study}

\begin{figure*}[htbp]
    \centering
    \includegraphics[width=\textwidth]{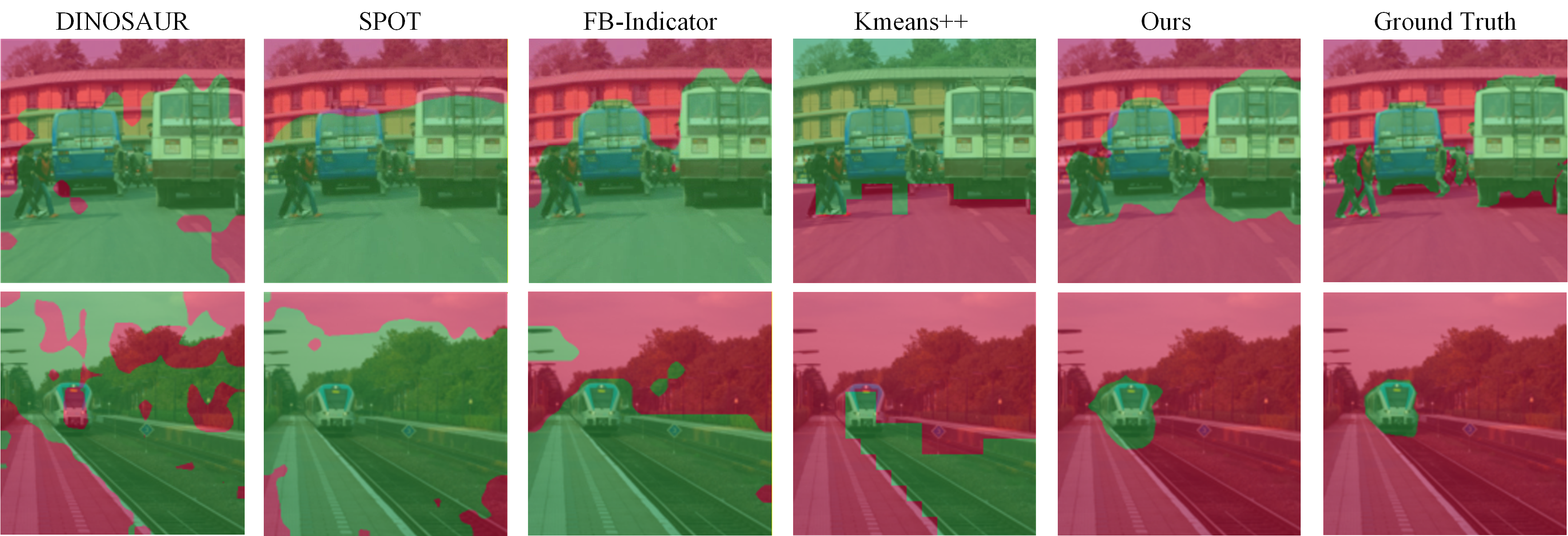}
    \caption{Visual comparison of \lyj{foreground–background separation}. \lyj{Green regions denote detected foreground; red regions indicate background. Our method achieves more accurate structural separation between foreground objects and background areas, generating masks that align closely with ground-truth annotations.} 
    }
    \label{fig:3}
\end{figure*}

\paragraph{Separation of Foreground and Background.}
We \yh{evaluate} the effectiveness of different methods in performing foreground–background separation. Table \ref{tab:fgbg} \lyj{reports} quantitative results, \lyj{with visual comparisons provided in Figure~\ref{fig:3}.} For DINOSAUR, SPOT, FB-Indicator, and \lyj{our} method, \lyj{we fix} the number of slots to two. For k-means++, we \lyj{similarly} use two clustering centroids. Following \cite{tian2025pay}, we adopt Mean Best Overlap (mBO) as the evaluation metric. In \lyj{our} experiment, ground-truth masks are  binarized into foreground (1) and background (0), \lyj{and predictions are matched to ground truth via} best-overlap matching strategy.


\begin{table}[htbp]
\setlength{\abovecaptionskip}{0.1cm}
\centering
\caption{Comparison of different methods \lyj{on} foreground and background separation.}
\label{tab:fgbg}
\begin{tabular*}{\linewidth}{@{\extracolsep{\fill}}lcc}
\toprule
Model & VOC & COCO \\
\midrule
DINOSAUR (2 slots) \cite{seitzer2022dinosaur}& 47.6 & 46.9 \\
SPOT (2 slots) \cite{kakogeorgiou2024spot}& 50.8 & 46.4 \\
FB-Indicator (2 slots) \cite{tian2025pay}& 52.6 & 48.4 \\
k-means++ \cite{arthur2006kmeans++}& 47.5 & 43.6 \\
\textbf{Ours} & \textbf{60.1} & \textbf{51.8} \\
\bottomrule
\end{tabular*}
\end{table}

\lyj{As shown in Table~\ref{tab:fgbg}, our method achieves the best performance on both COCO and PASCAL datasets, demonstrating the advantage of our clustering-based initialization strategy in producing accurate and stable foreground–background separation. In contrast, while baselines such as DINOSAUR and SPOT can partition scenes into two regions, they often fail to precisely separate foreground from background. Visual results in Figure~\ref{fig:3} further illustrate that these methods frequently include background areas in foreground predictions, whereas our approach cleanly distinguishes object regions. Although FB-Indicator and standalone k-means++ also separate foreground and background to some extent, our method consistently outperforms them in both qualitative and quantitative evaluations. More visualization results are included in the supplementary material.}

\paragraph{Unsupervised salient object detection.}
We further compare our method with state-of-the-art unsupervised salient object detection techniques \cite{wang2023tokencut,wang2022umnet,shin2022selfmask,wang2023cutler,simeoni2023localization,yasarla20243sd}. \lyj{Table \ref{tab:performance} summarizes results on the PASCAL-S dataset \cite{li2014pascals}, evaluated using pixel accuracy (Acc), intersection-over-union (IoU), and $F_{\beta}$ score as in \cite{simeoni2023localization}. Our method achieves competitive or superior performance across all metrics, indicating its effectiveness in capturing foreground saliency and establishing a reliable basis for subsequent object-level decomposition.}
\begin{table}[htbp]
\setlength{\abovecaptionskip}{0.1cm}
\centering
\caption{Comparison of \lyj{different methods on} object localization.}
\label{tab:performance}
\begin{tabular*}{0.8\linewidth}{@{\extracolsep{\fill}}lccc}
\toprule
Model & Acc & IoU & $F_{\beta}$ \\
\midrule
TokenCut \cite{wang2023tokencut} & 89.7 & 67.5 & 0.782 \\
UMNET \cite{wang2022umnet} & 89.6 & 63.9 & 0.771 \\
SelfMask \cite{shin2022selfmask} & 90.1 & 69.6 & 0.806 \\
CutLER \cite{wang2023cutler} & 89.1 & 67.7 & 0.764 \\
Found \cite{simeoni2023localization} & 91.3 & 71.5 & \textbf{0.815} \\
3SD \cite{yasarla20243sd} & 90.6 & 69.1 & 0.791 \\
\textbf{Ours} & \textbf{91.7} & \textbf{72.3} & 0.809 \\
\bottomrule
\end{tabular*}
\end{table}

\begin{figure}[htbp]
\setlength{\abovecaptionskip}{0.02cm}
    \centering
    \includegraphics[width=0.45\textwidth]{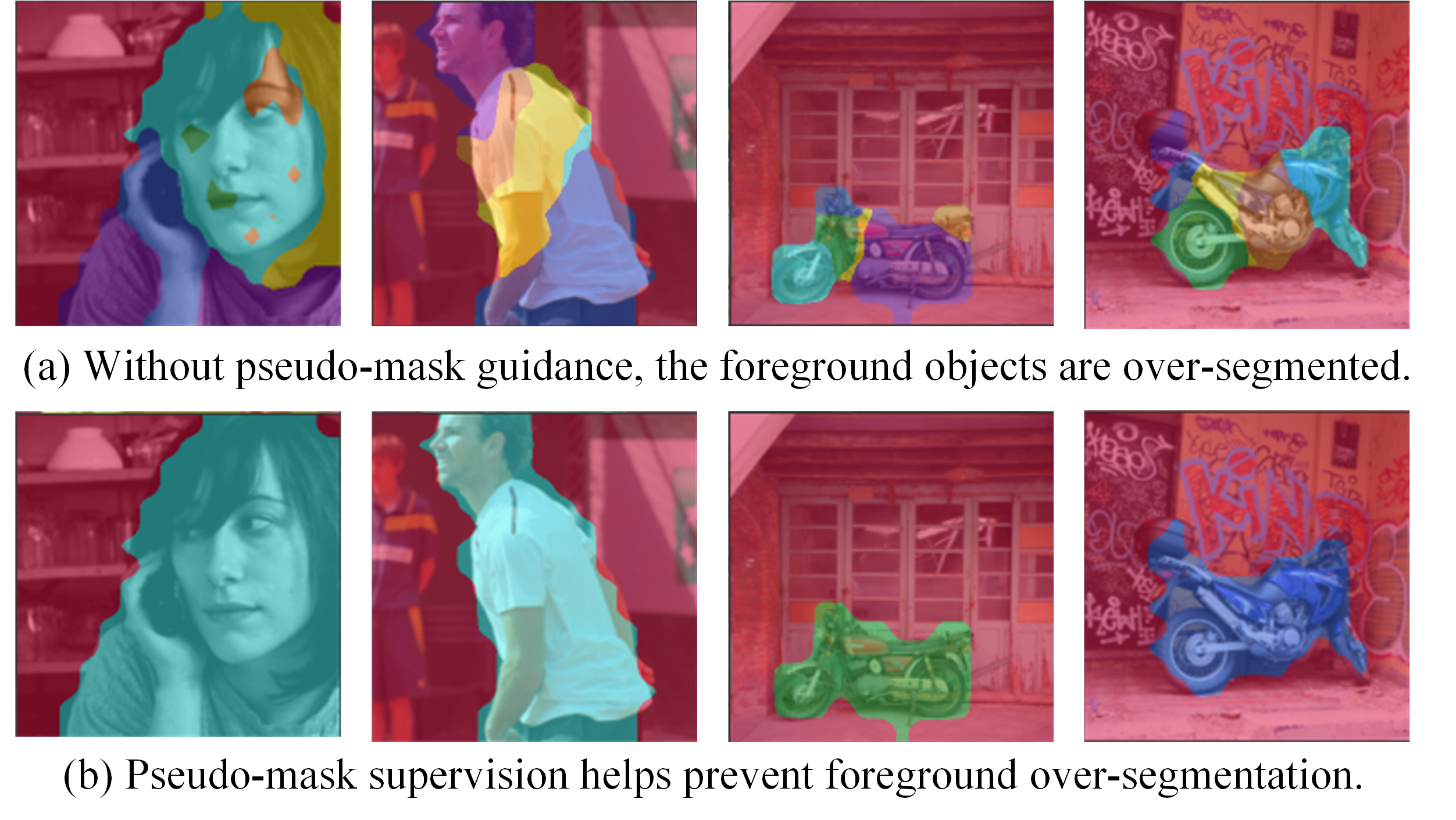}
    \caption{Visualization of results with and without pseudo-mask supervision. (a) Without pseudo-mask supervision; (b) With pseudo-mask supervision. Using pseudo-mask supervision can effectively reduce over-segmentation.}
    \label{fig:4}
\end{figure}

\paragraph{\lyj{Impact} of Pseudo-Mask Guidance.}
\lyj{We conduct an ablation study to evaluate the contribution of pseudo-mask guidance by removing the corresponding loss term during training on the COCO dataset. As shown in Table \ref{tab:3}, the variant $w/o~\mathcal{L}_{\mathrm{BCE}}$ exhibits significant performance degradation. More critically, qualitative results in Figure \ref{fig:4} reveal that without pseudo-mask guidance, the model becomes prone to over-segmenting foreground objects.}
This occurs because the slots compete to parse the foreground objects independently. When the number of objects in the foreground is small, this competition causes the slots to partition a single object into multiple fragments, thereby resulting in over-segmentation. \lyj{Additional analysis is provided} in the supplemental material.

\vspace{-0.3cm}

\paragraph{Effect of Different Slot Numbers.}

The number of slots influences the model’s \lyj{capacity} to discover and represent multiple objects \lyj{in complex scenes}. To \lyj{evaluate this effect}, we conduct experiments using \lyj{varying} slot configurations. \lyj{As summarized in} Table \ref{tab:4}, \lyj{optimal performance on COCO and VOC datasets is achieved with 7 and 6 slots respectively, indicating that dataset characteristics influence the ideal slot configuration.} 

\begin{table}[t]
\setlength{\abovecaptionskip}{0.1cm}
\centering
\caption{The effect of pseudo-mask guidance.}
\label{tab:3}
\begin{tabular}{lccc}
\toprule
Loss term & mIoU & mBO$^{i}$ & mBO$^{c}$ \\
\hline
$w/o~\mathcal{L}_{\mathrm{BCE}}$ & 25.2 & 26.8 & 30.5 \\
$w/~\mathcal{L}_{\mathrm{BCE}}$ & \textbf{34.1} & \textbf{36.5} & \textbf{43.9} \\
\bottomrule
\end{tabular}
\end{table}
\vspace{-0.4cm}

\begin{table}[htbp]
\setlength{\abovecaptionskip}{0.1cm}
\centering
\caption{Effects of different $K$ values.}
\label{tab:4}
\begin{tabular}{lcccc}
\toprule
Dataset & $K$ & mIoU & mBO$^{i}$ & mBO$^{c}$ \\
\hline
\multirow{4}{*}{COCO} 
 & 5  & 30.3 & 31.6 & 35.4 \\
 & 6  & 32.6 & 34.5 & 41.9 \\
 & 7  & \textbf{34.1} & 36.5 & \textbf{43.9} \\
 & 8  & 33.7 & \textbf{36.9} & 43.7 \\
 & 9& 32.8& 34.9&40.3\\
\hline
\multirow{4}{*}{VOC}  
 & 5  & 46.6 & 50.3 & 53.5 \\
 & 6  & \textbf{49.5} & \textbf{50.2} & \textbf{57.3} \\
 & 7  & 48.9 & 49.3 & 52.5\\
 & 8 & 46.9 & 47.3& 51.8\\
 & 9& 45.2& 46.9&49.1\\
\bottomrule
\end{tabular}
\end{table}
\vspace{-0.7cm}



\subsection{Downstream Applications}

\lyj{To further evaluate the effectiveness and generalization capability of the learned slot representations, we investigate two downstream tasks: object localization \cite{simeoni2023localization} and zero-shot object discovery \cite{singh2025glass}.}

\begin{table}[htbp]
\setlength{\abovecaptionskip}{0.1cm}
\centering
\caption{\lyj{Comparison on the object localization task.}}
\label{tab:voc_coco}
\begin{tabular*}{0.8\linewidth}{@{\extracolsep{\fill}}lcc}
\toprule
Model & VOC & COCO \\
\midrule
DINOSAUR \cite{seitzer2022dinosaur} & 0.047 & 0.071 \\
StableLSD \cite{lsd} & 0.045 & 0.067 \\
SPOT \cite{kakogeorgiou2024spot} & 0.042 & 0.064 \\
FB-Indicator \cite{tian2025pay} & 0.044 & 0.066 \\
\textbf{Ours} & \textbf{0.038} & \textbf{0.061} \\
\bottomrule
\end{tabular*}
\end{table}

{\noindent\bf Object Localization.} This task involves predicting an object's bounding box center from its corresponding slot representation. Following \cite{seitzer2022dinosaur}, we evaluate on COCO and PASCAL VOC datasets using mean squared error (MSE) as the evaluation metric. As shown in Table \ref{tab:voc_coco}, our method achieves superior localization accuracy. The results confirm that our learned slot representations encode spatially coherent object information, enabling more precise position estimation through effective instance-level representations.

\begin{table}[htbp]
\setlength{\abovecaptionskip}{0.1cm}
\centering
\caption{\lyj{Comparison on the zero-shot object discovery task.}}
\label{tab:clevr_obj}
\resizebox{\linewidth}{!}{%
\begin{tabular}{lcccc}
\toprule
\multirow{2}{*}{Model} & \multicolumn{2}{c}{CLEVRTex} & \multicolumn{2}{c}{Obj365} \\
\cmidrule(lr){2-3} \cmidrule(lr){4-5}
 & $\mathrm{mIoU}$ & $\mathrm{mBO}^{i}$ & $\mathrm{mIoU}$ & $\mathrm{mBO}^{i}$ \\
\midrule
DINOSAUR \cite{seitzer2022dinosaur} & 30.2 & 35.1 & 16.2 & 18.9 \\
StableLSD \cite{lsd} & 24.0 & 27.6 & 14.8 & 16.9 \\
SPOT \cite{kakogeorgiou2024spot} & 39.5 & 43.7 & 18.0 & 20.7 \\
FB-Indicator \cite{tian2025pay} & 39.7 & 44.2 & 18.2 & 20.8 \\
\textbf{Ours} & \textbf{40.6} & \textbf{44.9} & \textbf{18.4} & \textbf{21.3} \\
\bottomrule
\end{tabular}%
}
\end{table}

{\noindent\bf Zero-Shot Object Discovery.} This task evaluates a model's ability to discover objects in unseen datasets. We train our model on COCO and evaluate its generalization performance on CLEVRTex \cite{karazija2021clevrtex} and Objects365 \cite{shao2019objects365} datasets. As shown in Table \ref{tab:clevr_obj}, our approach outperforms existing methods. By learning foreground-biased slot representations, our method produces more accurate instance-level segmentation masks, demonstrating strong generalization capability for zero-shot object discovery.

%% file: sec/5_conclusion.tex
\section{Conclusion}
In this work, we introduced Foreground-Aware Slot Attention (FASA), a two-stage framework designed to enhance unsupervised scene decomposition \lyj{through explicit foreground–background separation}. \lyj{By integrating dual-slot competition with clustering-based initialization,} FASA achieves structured scene representations while mitigating background interference. The \lyj{proposed} masked slot attention mechanism enables fine-grained foreground object discovery, guided by a pseudo-mask to prevent over-segmentation. Extensive experiments on both synthetic and real-world datasets confirm that FASA \lyj{consistently} outperforms existing methods.

{\noindent\bf Limitations and Future Work.} Despite its strong performance, our approach shares certain limitations with existing object-centric methods. It struggles in crowded scenes containing numerous similarly appearing objects, where instance-level discrimination remains challenging. Additionally, the model shows limited sensitivity to small, spatially isolated objects, which may be overlooked or absorbed into background representations. We provide an analysis of these failure cases in the supplementary material. Future work will focus on enhancing discrimination in dense scenes and improving small-object awareness.


%% file: sec/X_suppl.tex
\clearpage
\setcounter{page}{1}
\setcounter{figure}{0}
\setcounter{table}{0}
\setcounter{equation}{0}
\maketitlesupplementary
\setcounter{section}{0}
\renewcommand\thesection{\Alph{section}}
\renewcommand{\thefigure}{S\arabic{figure}}
\renewcommand{\thetable}{S\arabic{table}}
\renewcommand{\theequation}{S\arabic{equation}}

\section{Implementation Details}
\label{sec:Implementation details}
\subsection{Slot Initialization}
In the first-stage \textit{foreground–background separation}, the encoder features are clustered into two groups, and the resulting cluster centers are used to initialize the two slots. To obtain a stable and well-separated initialization, we employ the K-Means++ strategy \cite{arthur2006kmeans++}, which randomly selects the first center and then iteratively samples subsequent centers with probabilities proportional to their squared distances from the existing centers. Subsequently, we apply the standard K-Means algorithm, in which each feature is assigned to its nearest cluster center, and the centers are iteratively updated as the mean of their assigned samples until convergence. The resulting cluster centers are then projected to the predefined slot dimension $D_{\text{slots}}$ via a linear transformation layer to ensure compatibility with the slot representations. In our implementation, we set $D_{\text{slots}} = 256$.

For the second-stage \textit{foreground decomposition}, the initial set of $K$ slots are randomly sampled from a Gaussian distribution parameterized by learnable mean $\mu \in \mathbb{R}^{D_{\text{slots}}}$ and log standard deviation $\log \sigma \in \mathbb{R}^{D_{\text{slots}}}$.

\subsection{Implementation of Slot Attention}
\label{sec:SA}
Our slot attention implementation is based on the formulation of Locatello et al. \cite{locatello2020object}. 
We begin by normalizing the $N$ input features and projecting them into the slot space via a two-layer MLP, resulting in features of dimension $D_{\text{slots}}$. The model initializes slot vectors by 
clustering (first stage) or sampling from a Gaussian distribution (second stage). During 
iterative updates, each step includes: (1) normalizing the slot representations, (2) 
performing attention to assign input features to slots, (3) updating slot states through a 
GRU \cite{cho2014gru}, and (4) applying a residual two-layer MLP. The query, key, and value projections all 
have dimensionality $D_{\text{slots}} = 256$ and omit bias terms. The MLP used in this 
process has a hidden dimension of $4 \times D_{\text{slots}}$.

\subsection{MLP Decoder}
\label{sec:mlp}
Our MLP-based decoder adopts the spatial broadcast mechanism from prior studies \cite{watters2019MLP}\cite{seitzer2022dinosaur}. Each slot is expanded into $N$ tokens that represent image patches, and augmented with learnable positional embeddings for spatial reference. These tokens are independently processed by a four-layer MLP with ReLU activations, producing both a reconstruction vector and an alpha (attention) map per slot. The alpha map identifies the spatial regions activated by each slot. The final feature reconstruction is generated by combining all slot reconstructions, where alpha maps act as weighting coefficients.

\subsection{MaskCut}

In practice, we construct the binary similarity matrix $W$ in Eq. 6 by 
applying a threshold of $\tau = 0.15$. Elements satisfying $W_{ij} < \tau$ are 
replaced with $1 \times 10^{-5}$, while those satisfying $W_{ij} \ge \tau$ are 
assigned a value of $1$.

MaskCut generate multiple object masks in an image by iteratively applying Normalized Cuts (NCut)  \cite{wang2023tokencut}  on the similarity matrix $W$. Once the bipartition is obtained via NCut at iteration $t$, we derive two disjoint 
patch groups and construct a foreground mask $M^{t}$. Specifically, 
$M^{t}_{i} = 1$ if $x_{i} > \mathrm{mean}(x^{t})$, indicating foreground, and 
$M^{t}_{i} = 0$ otherwise, indicating background. The vector $x^{t}$ is the 
eigenvector associated with the second smallest eigenvalue $\lambda$ as illustrated in Section 3.3. 

We reverse the partitioning of the foreground and background, i.e., 
$M^{t}_{i} = 1 - M^{t}_{i}$, if the foreground set should contain more than two 
of the four corners. We then mask out the discovered foreground using the 
foreground mask:
\begin{equation}
W_{ij}^{t+1}
=
\frac{
\left( K_i \prod_{s=1}^{t} \hat{M}_i^{s} \right)
\left( K_j \prod_{s=1}^{t} \hat{M}_j^{s} \right)
}{
\lVert K_i \rVert_2^{2} \, \lVert K_j \rVert_2^{2}
}
\end{equation}
where 
$\hat{M}_{i}^{s} = 1 - M_{i}^{s}$, $K_i$ is the "key" feature of patch $i$.

By iteratively performing NCut \cite{wang2023tokencut} and removing the discovered foreground regions from the similarity matrix $W$ at each iteration, multiple object masks can be extracted from a single image. Following CutLER \cite{wang2023cutler}, we set the number of iterations to $n=3$.

\subsection{ Training Details}
\label{sec:fasa}
Our FASA (Foreground-Aware Slot Attention) framework employs a ViT-S/14 encoder initialized with DINOv2~\cite{oquab2023dinov2}, 
omitting the final layer normalization. In the MLP-based decoder, the hidden layer size is 
set to 2048. Across all experiments, the slot attention module is iterated three times, 
with a slot dimension of $256$ and an MLP hidden dimension of $1024$. We use the Adam 
optimizer with parameters $\beta_{1}=0.9$ and $\beta_{2}=0.999$, no weight decay, and a 
batch size of $64$. 

Our training pipeline is divided into two stages. In stage 1, we optimize the foreground–background separation module illustrated in the upper portion of Fig. 1. Once trained, this module is frozen. In stage 2, we train the foreground decomposition module shown in the lower portion of Fig. 1. For both stages, the learning rate follows a linear warm-up from $0$ to its peak value over $10{,}000$ 
iterations, followed by cosine annealing decay. For experiments on COCO  \cite{lin2014coco} and PASCAL VOC \cite{everingham2010voc}, the 
peak learning rate is set to $1\times10^{-4}$ and the minimum value to $1\times10^{-7}$. 
For MOVi-C \cite{greff2022movic} experiments, the peak learning rate is $2\times10^{-4}$ and the minimum value 
is $1\times10^{-5}$.

\section{More Experimental Results}

\subsection{Comparison between Transformer Decoder and MLP Decoder}

\begin{table}[h]
\centering
\caption{Comparison between different decoders.}
\label{tab:decoder_eval}
\begin{tabular*}{\linewidth}{@{\extracolsep{\fill}} l c c c}
\toprule
Decoder & mIoU & mBo$^{i}$ & mBo$^{c}$ \\
\midrule
Transformer & 32.3 & 34.1 & 39.7 \\
MLP         & \textbf{34.1} & \textbf{36.5} & \textbf{43.9} \\
\bottomrule
\end{tabular*}
\end{table}

In this study, we adopt an MLP decoder, as it is more effective at facilitating instance-level scene decomposition \cite{seitzer2022dinosaur}\cite{singh2025glass}. To further examine the influence of decoder design, we compare the MLP decoder with a Transformer-based decoder on the COCO dataset. The quantitative results are summarized in Table~\ref{tab:decoder_eval}. We observe that the MLP decoder outperforms its Transformer-based counterpart, highlighting its advantage within our framework.
\subsection{Comparison with Different Pre-trained Image Features}

\begin{table}[htbp]
\centering
\caption{Evaluation with various pre-trained encoders on COCO.}
\label{tab:encoder_eval}
\begin{tabularx}{\linewidth}{lYYY}
\toprule
Encoder & mIoU & mBO$^{i}$ & mBO$^{c}$ \\
\midrule
DINOv1 \cite{caron2021dino}   & 32.5 & 34.2 & 42.8 \\
MOCO-v3 \cite{chen2021MOCO} & 30.6 & 31.5 & 40.1 \\
MAE \cite{he2022MAE}      & 31.9 & 33.7 & 41.2 \\
\textbf{Ours}  & \textbf{34.1} & \textbf{36.5} & \textbf{43.9} \\
\bottomrule
\end{tabularx}
\end{table}

In our study, we adopt DINOv2 \cite{oquab2023dinov2} as the image encoder. To further understand how different pre-training strategies affect the performance of our framework, we additionally conduct experiments on the COCO dataset using features extracted from several alternative self-supervised models, including DINOv1 \cite{caron2021dino}, MOCO-v3 \cite{chen2021MOCO}, and MAE \cite{he2022MAE}. As reported in Table \ref{tab:encoder_eval}, our method consistently achieves the best results when DINOv2 is used, outperforming the other pre-trained encoders by a clear margin. Based on this observation, we choose DINOv2 as the backbone for all subsequent experiments, given its superior representational quality and strong compatibility with our object discovery framework.

\subsection{Comparison with FG-ARI Scores}

\begin{table}[htbp]
\centering
\caption{Comparison between OCL methods for the FG-ARI metric.}
\label{tab:ocl_fgar}
\begin{tabularx}{\linewidth}{lYY}
\toprule
\textbf{Model} & \textbf{VOC} & \textbf{COCO} \\
\midrule
SA \cite{locatello2020object}          & 12.3 & 21.4 \\
SLATE \cite{singh2021slate}       & 15.6 & 32.5 \\
DINOSAUR \cite{seitzer2022dinosaur}     & 24.8 & 34.3 \\
StableLSD \cite{lsd}   &  8.7 & 35.0 \\
SPOT \cite{kakogeorgiou2024spot}        & 19.9 & 37.8 \\
FB-Indicator \cite{tian2025pay} & 25.0 & 38.1 \\
\textbf{Ours} & \textbf{25.6} & \textbf{38.7} \\
\bottomrule
\end{tabularx}
\end{table}

FG-ARI, a foreground-focused variant of the adjusted Rand index (ARI) \cite{rand1971ARI}, measures clustering similarity in a permutation-invariant manner while evaluating only the foreground regions of an image. FG-ARI has been a common metric in assessing predicted object masks against ground-truth segmentation in previous works\cite{kakogeorgiou2024spot}\cite{engelcke2019genesis}\cite{karazija2021clevrtex}. However, FG-ARI can be unreliable because it disregards background pixels and does not adequately reflect the structural accuracy of predicted masks \cite{kakogeorgiou2024spot}\cite{singh2025glass}. We also report FG-ARI results in Table \ref{tab:ocl_fgar}  for completeness and comparability with previous studies.  The results show that our method attains the highest performance, which we attribute to its ability to accurately capture and isolate foreground objects.

\subsection{Analysis of Loss Weight $\lambda$}


\begin{table}[t]
\centering
\caption{Effect of the loss weight $\lambda$ on COCO.}
\begin{tabular*}{0.8\linewidth}{@{\extracolsep{\fill}} c c c c}
\toprule
$\lambda$ & mIoU & mBo$^{i}$ & mBo$^{c}$ \\
\midrule
1     & 26.3 & 28.3 & 37.9 \\
0.5   & 29.2 & 31.1 & 38.7 \\
0.05  & \textbf{34.1} & \textbf{36.5} & \textbf{43.9} \\
0.005 & 30.5 & 32.6 & 39.2 \\
0     & 25.2 & 26.8 & 30.5 \\
\bottomrule
\end{tabular*}
\label{tab:lambda_ablation}
\end{table}

To evaluate the impact of the parameter $\lambda$ in Eq. 9, we conduct an ablation study by varying $\lambda$ across several settings on the COCO dataset. The experimental results are presented in Table \ref{tab:lambda_ablation}. The model achieves its optimal performance at $\lambda = 0.05$. As $\lambda$ decreases, the reconstruction loss $L_{\text{REC}}$ begins to dominate the training objective, leading to over-segmentation of foreground objects. In contrast, an excessively large $\lambda$ overweights the $L_{\text{BCE}}$ term, causing the number of foreground slots to collapse to the number of objects indicated by the pseudo mask (i.e., $n = 3$). This restricts the model's ability to discover additional objects and ultimately results in under-segmentation.

\subsection{Object Localization}

For the object localization task, we employ a single-layer linear model for predicting object positions. 
We follow the approach in \cite{seitzer2022dinosaur} to associate labels with their corresponding slots. Optimization is performed using AdamW \cite{loshchilov2017ADAMW} with a fixed learning rate of $3 \times 10^{-4}$. 
Image coordinates are normalized to the range $[0,1]$ by dividing by the image dimensions.

\begin{figure}[t]
    \centering
    \includegraphics[width=0.45\textwidth]{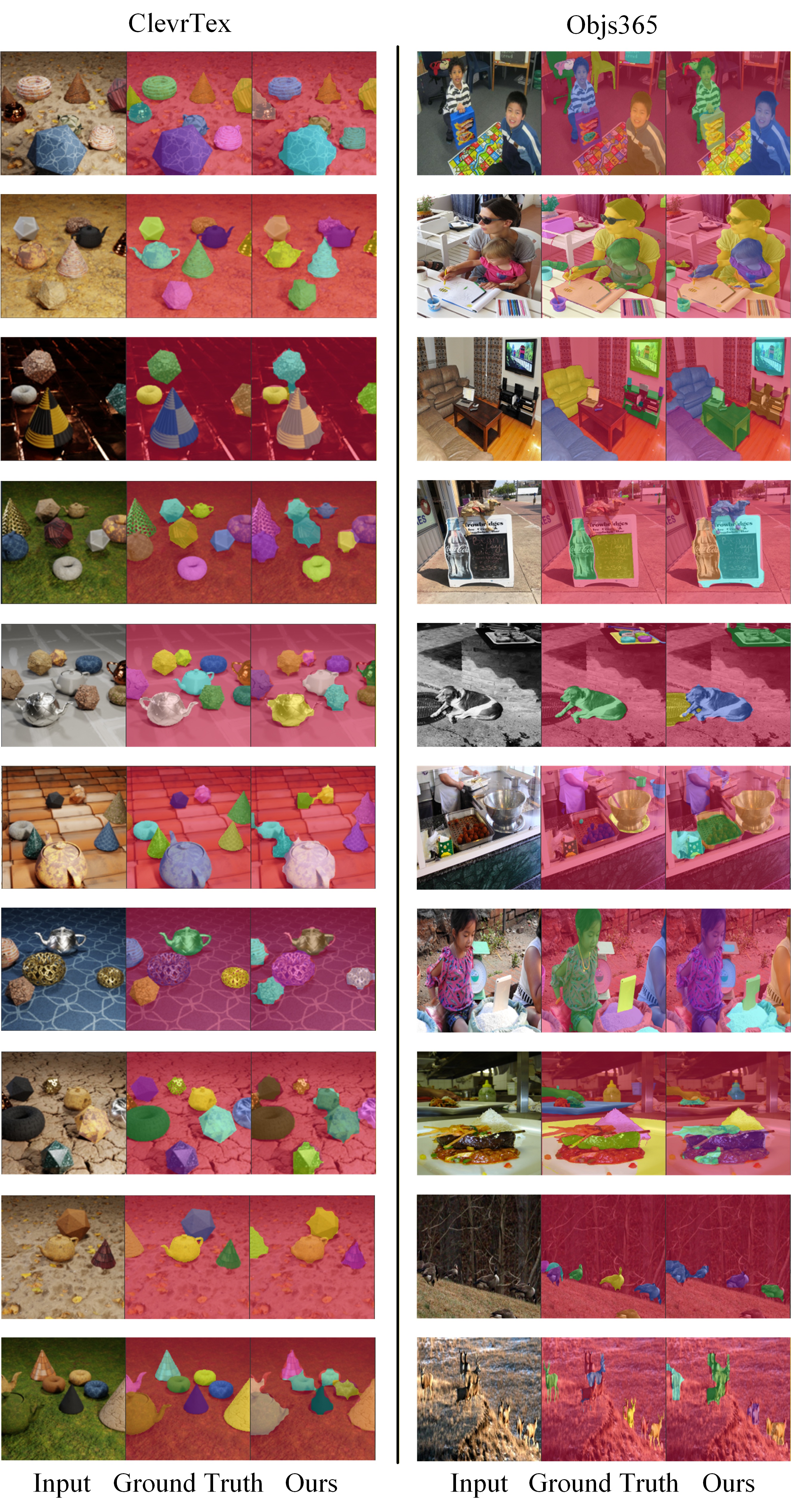}
    \caption{Visualization of zero-shot object discovery. Our approach exhibits strong generalization capabilities across both synthetic datasets and challenging real-world image datasets, maintaining robust performance even on data that significantly deviates from the training distribution.}
    \label{fig:zeroshot}
\end{figure}

\begin{figure}[t]
    \centering
    \includegraphics[width=0.45\textwidth]{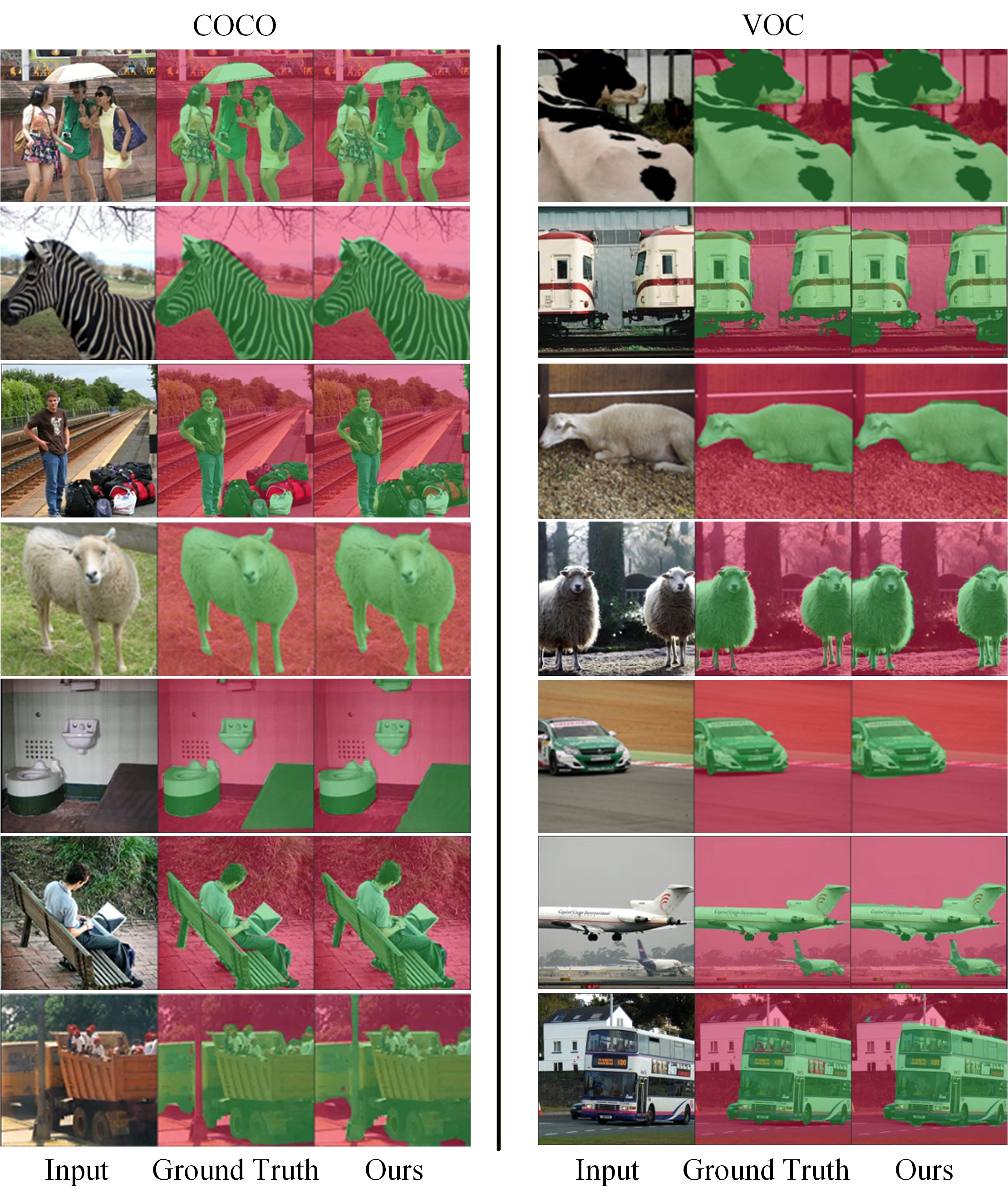}
    \caption{Visualization of foreground-background separation. In the figure, red denotes the background, and green denotes the foreground. Our method accurately detects the objects in the foreground and produces results that are close to the ground truth.}
    \label{fig:supfb}
\end{figure}

\begin{figure*}[htbp]
    \centering
    \includegraphics[width=0.8\textwidth]{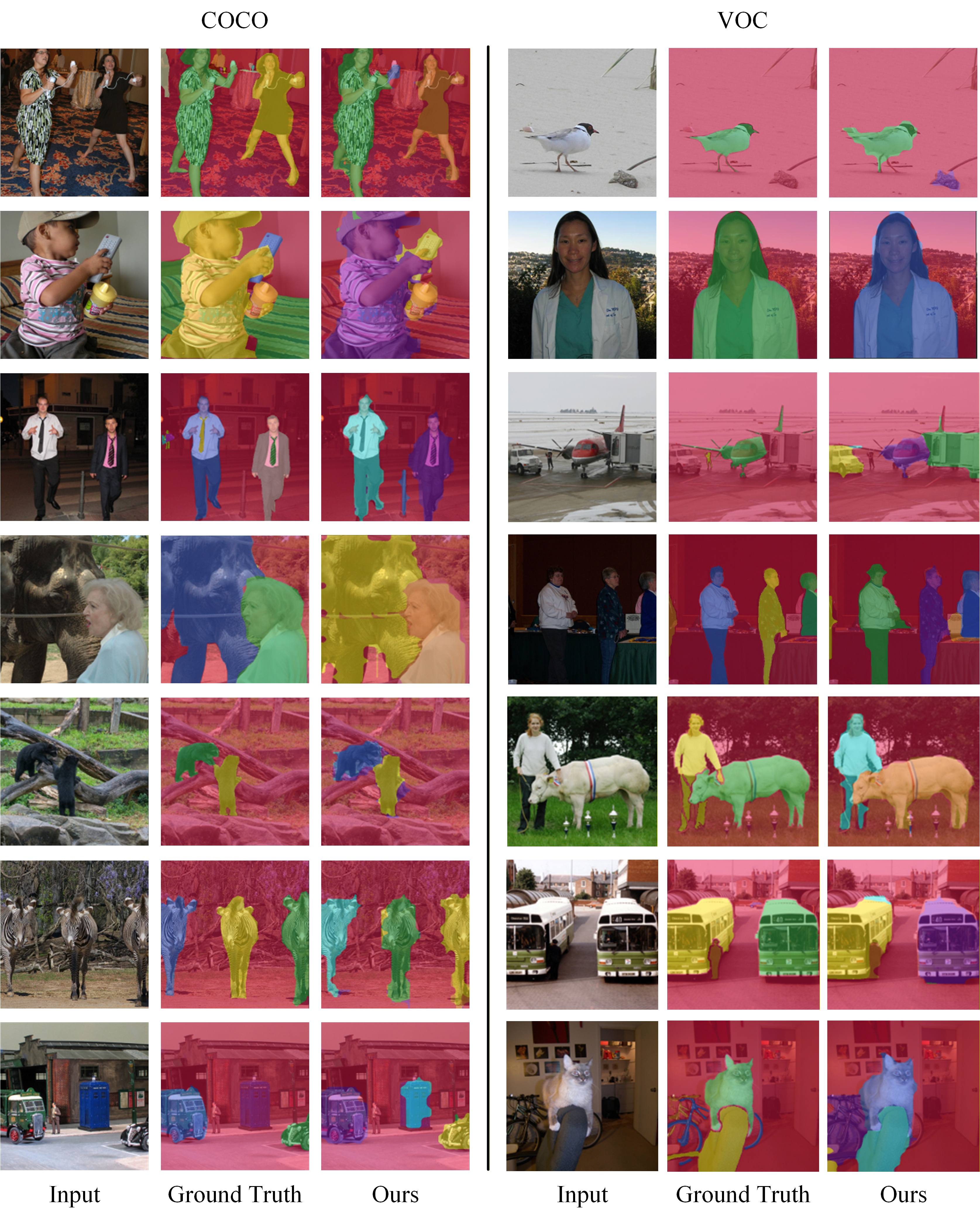}
    \caption{Visualization of object discovery. The red regions represent the background, whereas the other colors denote different foreground objects, each assigned to a separate slot. Our method is capable of producing instance-level object discovery results, yielding outputs that closely match the ground-truth annotations.}
    \label{fig:sup_obj_disc}
\end{figure*}

\begin{figure*}[htbp]
    \centering
    \includegraphics[width=0.8\textwidth]{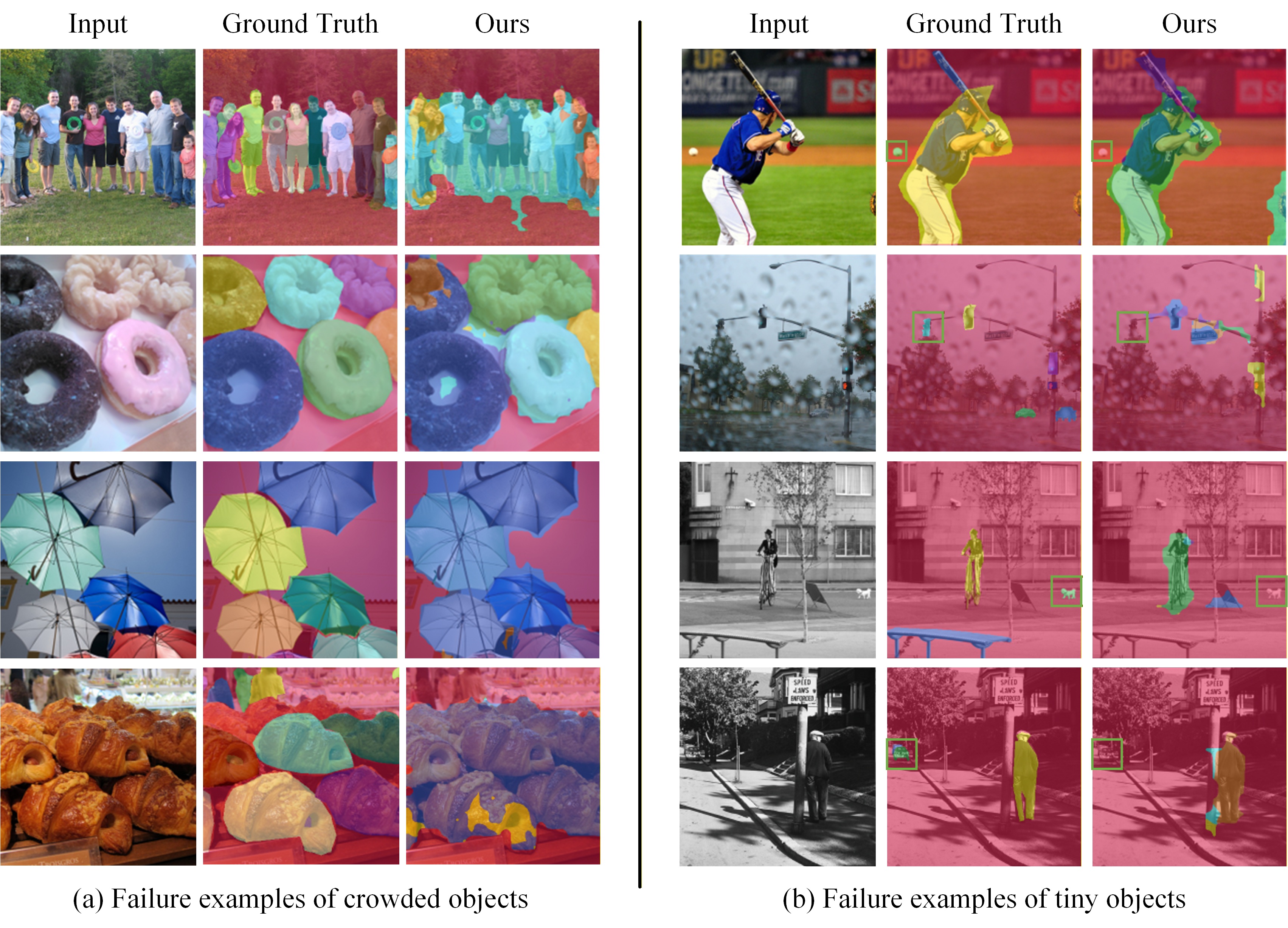}
    \caption{Failure examples. Our method struggles to achieve instance-level segmentation in highly crowded scenes. In addition, small objects may sometimes be merged into the background, as indicated by the green boxes.}
    \label{fig:fail}
\end{figure*}

\subsection{More Details on Zero-Shot Object Discovery Task}
For the zero-shot object discovery task, we train our model on COCO and evaluate its generalization ability on the test sets of the CLEVRTex \cite{karazija2021clevrtex} and Objects365 \cite{shao2019objects365} datasets. CLEVRTex is a texture-rich synthetic dataset designed for unsupervised multi-object segmentation. It incorporates realistic textures, materials, and lighting variations, yielding scenes that better approximate real-world visual complexity. In  CLEVRTex, each image contains 3–10 objects, and we use 10 slots for object discovery. Objects365 is a large-scale dataset for generic object detection, consisting of approximately 600K images and 10 million annotated bounding boxes across 365 categories. It provides highly diverse real-world scenes, making it a strong benchmark for assessing cross-dataset generalization. Since ground-truth masks are not available, we obtain the masks for Objects365 using SAMv2 \cite{ravi2024samV2}, and we use 7 slots for evaluation on this dataset.


Fig. \ref{fig:zeroshot} presents zero-shot object discovery results of our method on the two datasets described above. These visualization results show that our approach exhibits strong generalization ability across datasets and scenes.

On the CLEVRTex dataset, our method adaptively handles diverse background variations and complex object compositions. Despite differences in background textures, lighting conditions, and object counts or spatial arrangements, our model consistently produces object regions that closely align with the ground truth. It accurately identifies and localizes the primary instances in each image, demonstrating robustness to previously unseen visual patterns.

On the Objects365 dataset, our method achieves strong performance in more challenging real-world settings. Even in natural images containing substantial noise, complex occlusions, and large variations in object scale, our model can still precisely differentiate individual instances and achieve reliable instance-level binding. This highlights the method’s adaptability and stability when applied to real-world scenes.

Overall, these results demonstrate that our method generalizes effectively to out-of-distribution data and maintains stable performance under diverse and challenging conditions, reflecting strong generalization capability, robustness, and practical applicability.



\subsection{More Visualization Results}

We provide additional visualization results to further validate the effectiveness of our method.

Fig. \ref{fig:supfb} presents the foreground object detection results of our approach on the COCO and VOC datasets. As illustrated in the figure, our method demonstrates strong robustness across diverse visual scenes. Even in images containing cluttered backgrounds, varying illumination conditions, or multiple interacting instances, the model is able to reliably distinguish foreground objects from their surrounding context. It not only identifies the primary objects with high accuracy but also preserves fine-grained structural details, producing sharp and precise object boundaries. These results further validate the effectiveness of our approach in handling real-world scenarios and underscore its capability for accurate foreground extraction in challenging environments.

Fig. \ref{fig:sup_obj_disc} shows the object discovery results of our method on the COCO and VOC datasets. From the figure, we can observe that our approach demonstrates strong robustness in complex visual environments. On the one hand, it effectively separates the foreground from the background, preventing cluttered or textured backgrounds from interfering with the slot assignment process. This allows foreground slots to concentrate more precisely on the salient foreground regions, thereby improving the overall quality of object discovery. On the other hand, our method successfully achieves instance-level binding of foreground objects. It is able to assign different objects belonging to the same semantic category to distinct slots, while simultaneously avoiding unnecessary over-segmentation of individual instances. These results highlight the capability of our approach to capture fine-grained object structures and maintain coherent instance representations even in challenging real-world scenes.

\subsection{Analysis of Failure Cases}


Fig. \ref{fig:fail} presents several failure cases of our method. As shown in Fig. \ref{fig:fail} (a), our approach struggles to achieve instance-level object discovery in highly crowded scenes—an issue commonly observed in slot attention–based models. This limitation can be attributed to several factors.
First, slots are global latent representations that do not explicitly encode object boundaries, geometric structures, or occlusion relationships. Consequently, when objects are densely packed or heavily occluded, the attention mechanism has difficulty assigning spatially continuous regions to distinct slots.
Second, our method is trained in a fully unsupervised setting without semantic or structural priors, making it challenging to distinguish adjacent objects with similar appearance or shape.
Finally, the information from multiple objects can become highly entangled on the feature map. This exacerbates the difficulty of separating individual instances and often results in under-segmentation or incorrect merging in crowded scenes.

As shown in Fig. \ref{fig:fail} (b), our method may fail to detect extremely small objects. This limitation primarily arises from the low visual saliency of such tiny instances: their appearance often contains only a few pixels with weak contrast against the background. As a result, these objects are easily absorbed into surrounding background regions during the foreground–background separation stage, causing the model to overlook them or fail to assign them to dedicated slots. This observation suggests that enhancing sensitivity to fine-scale details could further improve the method’s performance on scenes containing tiny objects.